%% file: root.tex
\documentclass[letterpaper, 10 pt, conference]{ieeeconf}  

\IEEEoverridecommandlockouts                              

\usepackage{graphics} 

\usepackage{mathptmx} 
\usepackage{times} 
\usepackage{amsmath} 
\usepackage{amssymb}  
\usepackage{graphicx}
\usepackage{makecell}
\usepackage{multirow}
\usepackage{tabularx}
\usepackage{float}
\usepackage[table,dvipsnames]{xcolor}
\input{tables/rankcolors}   
\usepackage[font=small]{caption}
\usepackage[hidelinks]{hyperref} 

\title{\LARGE \bf
VeloBins: Learning Velocity and Its Uncertainty via Bins and Error-Conditioned Gaussian Labels for Aerial Inertial Odometry
}

\author{Maulana Bisyir Azhari, Seungwook Lee, Donghun Han, Sung Jun Park, and David Hyunchul Shim
\thanks{This research was financially supported by the Institute of Civil Military Technology Cooperation funded by the Defense Acquisition Program Administration, and the Ministry of Trade, Industry, and Energy of Korean Government under Grant UM22206RD3.}
\thanks{All authors are with the Unmanned Systems Research Group (USRG), School of Electrical Engineering, Korea Advanced Institute of Science and Technology (KAIST), Daejeon 34141, South Korea. {\tt\small \{mbazhari, seungwook1024, donghun.han, sjpark, hcshim\}@kaist.ac.kr}}
}

\begin{document}

\maketitle
\thispagestyle{empty}
\pagestyle{empty}

\begin{abstract}

Inertial odometry (IO) is critical for aerial robots, where aggressive maneuvers and poor lighting degrade visual sensors.
Recent learning-based IO methods improve traditional integration-based approaches by learning motion priors from IMU and platform-specific sensors, then fusing the predictions within an extended Kalman filter.
However, learning velocity through regression is difficult, while jointly estimating uncertainty with a separate decoder and negative log-likelihood (NLL) loss further complicates training and can lead to over-confident estimates.
We introduce \textbf{VeloBins}, which reformulates velocity regression as classification over discretized velocity bins.
We decode both the velocity from the bin distribution's expectation and the uncertainty from its variance, removing the need for a separate uncertainty decoder.
We further supervise the uncertainty explicitly using an error-conditioned Gaussian label centered at the ground-truth velocity, with a standard deviation set to the velocity error.
We evaluate VeloBins on four aerial datasets, ranging from free-form aggressive flights and a 27\,g nano-quadrotor to drone racing at over 21~m/s.
VeloBins achieves the lowest average errors on all four datasets, reducing velocity, relative trajectory, and absolute trajectory errors by 3--27\%, 8--40\%, and 6--53\%, respectively, compared with the strongest baseline.
Notably, the proposed supervision achieves the lowest NLL and best filter consistency despite never optimizing an NLL loss.
The code will be released upon acceptance.

\end{abstract}

{\raggedright\small\noindent\textbf{Supplementary Video:} \url{https://youtu.be/QkZY0So3myw}\par}

\section{Introduction}
Inertial Odometry (IO) estimates a robot's trajectory using an Inertial Measurement Unit (IMU), a sensor that is ubiquitous, cost-effective, and lightweight\cite{chen2024deepio_survey}.
Unlike vision, it is unaffected by poor lighting, motion blur, or textureless scenes\cite{herath2020ronin, liu2020tlio, buchanan2022learning_io_legged, qiu2025airio}.
These properties make IO critical for aerial robots, which must estimate their state through aggressive maneuvers and low light, both of which degrade visual sensing\cite{qiu2025airio,cioffi2023imo}.
However, traditional IO directly integrates raw IMU measurements, where the inherent noise and biases accumulate unboundedly. This ``curse of drift''\cite{chen2018ionet} renders the estimate unusable within seconds.

\begin{figure}[t!]
\begin{center}
\vspace{8pt}
\includegraphics[width=0.9\columnwidth]{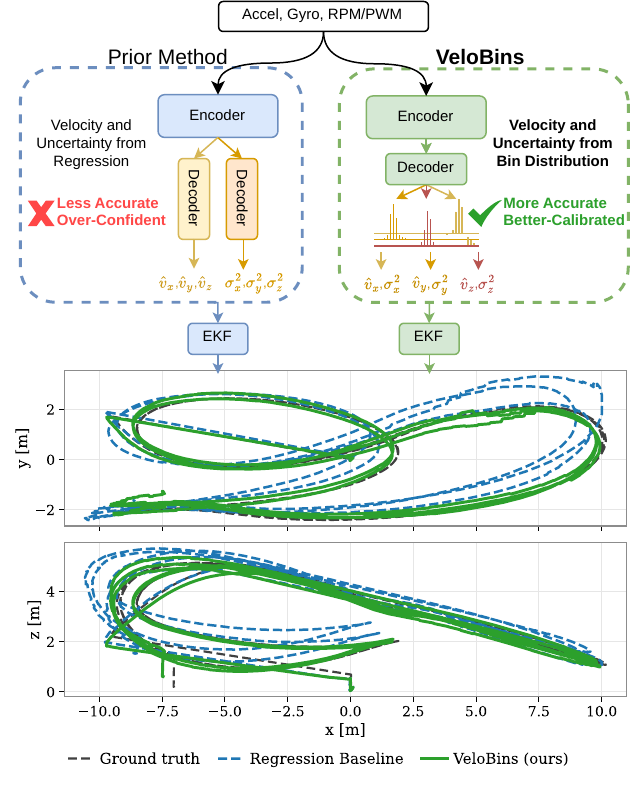}
\vspace{-8pt}
\caption{Top: Prior IO method~\cite{cui2026aiio} regresses the velocity and its uncertainty from two separate decoders, supervised by an NLL loss. \textbf{VeloBins} predicts a distribution over velocity bins, decoding the velocity as its expectation and the uncertainty as its variance, supervised with our error-conditioned Gaussian labels, yielding more accurate and better-calibrated estimates. Bottom: EKF-fused trajectories on a 21~m/s racing flight, where VeloBins follows the ground truth closely while the regression baseline drifts.}
\vspace{-22pt}
\label{fig:teaser}
\end{center}
\end{figure}

Learning-based IO overcomes this by learning a motion prior that maps a window of IMU data to a displacement or velocity\cite{herath2020ronin,liu2020tlio,chen2018ionet}.
The prediction is fused with IMU propagation in an extended Kalman filter (EKF), where each update is weighted by a per-sample uncertainty that the network also predicts\cite{liu2020tlio,qiu2025airio}.
Yet this formulation has two limitations.
First, existing IO methods regress the velocity or displacement directly\cite{herath2020ronin,qiu2025airio,cui2026aiio}, and scalar regression losses can be difficult to train\cite{imani2018histloss,zhang2023ordinal}.
Velocity regression is even harder for aerial robots, whose highly dynamic and non-linear flight differs from the pedestrian motion on which most learned priors are trained\cite{qiu2025airio,cioffi2023imo}.
Second, the measurement covariance comes from a separate uncertainty decoder trained by a negative log-likelihood (NLL) loss\cite{liu2020tlio,qiu2025airio,cui2026aiio}, an objective that is difficult to optimize and may yield over-confident variances\cite{seitzer2022pitfalls} that degrade filter consistency\cite{barshalom2001estimation}.

Both stem from representing the velocity and its uncertainty as point estimates, so we explore reformulating velocity regression as classification over velocity bins (Fig.~\ref{fig:teaser}).
Such bin-based formulations outperform direct regression in computer vision tasks such as monocular depth estimation\cite{Fu_2018_dorn_depth,bhat2021adabins} and human pose estimation\cite{li2022simcc,lu2024rtmo}.
The same reformulation improves value-based reinforcement learning\cite{bellemare2017distributional,farebrother2024stopregressing}.
The advantage is attributed to the classification objective, which provides more stable gradients and higher-entropy feature representations than a scalar error loss\cite{imani2018histloss,zhang2023ordinal}.
In this study, we investigate whether the same advantage holds for inertial odometry, where a network must predict the body-frame velocity of a dynamic platform.

Additionally, the bin distribution potentially offers the velocity uncertainty through its variance, which removes the need for a separate uncertainty decoder.
However, a velocity target alone supervises the distribution only through its expectation\cite{bhat2021adabins,li2024binsformer}, leaving the variance unconstrained and therefore the uncertainty measure unreliable.
Prior work supervises the distribution explicitly with a maximum-likelihood objective that again needs a separate decoder\cite{lu2024rtmo}, or with a Gaussian label of fixed standard deviation that cannot reflect the per-sample uncertainty\cite{imani2018histloss,gao2018dldlv2,jiang2023rtmpose}.
We therefore condition the label width on the current velocity error, so the supervision reflects the per-sample uncertainty.

To summarize, our contributions are:
\begin{itemize}
    \item \textbf{VeloBins}, a novel inertial odometry framework that predicts velocity as a distribution over discretized bins, decoding both the velocity and its uncertainty from the same distribution.
    \item An \textbf{error-conditioned Gaussian label} whose standard deviation is the current velocity error, supervising the velocity uncertainty explicitly without a separate uncertainty decoder or NLL loss.
    \item An extensive evaluation on four aerial datasets, from a 27\,g nano-quadrotor to 21~m/s drone racing, where VeloBins achieves the lowest average errors against five baselines and better filter consistency than the regression and likelihood-supervised variants.
\end{itemize}

\section{Related Work}

\subsection{Inertial Odometry}
Traditional IO estimates the trajectory by integrating IMU measurements through a strapdown process\cite{groves2013principles}, where the gyroscope is integrated once for the orientation and the accelerometer twice for the velocity and the position.
The bias and noise of a low-cost IMU are amplified at every integration stage, rendering long-term estimates unreliable\cite{chen2018ionet}.
Classical systems suppress this drift with motion constraints such as zero-velocity conditions\cite{brossard2019_rins_w}, which a UAV in continuous flight does not offer.

Learning-based IO instead learns a motion prior directly from data.
Seminal works such as IONet\cite{chen2018ionet} and RoNIN\cite{herath2020ronin} used temporal networks to regress motion from IMU windows and sharply reduce long-term drift.
Because a raw network estimate is still noisy, a parallel line fuses it with IMU propagation in a probabilistic filter.
TLIO\cite{liu2020tlio} feeds the network output \emph{and its predicted uncertainty} into an EKF.
AI-IMU\cite{brossard2020ai_imu_dr} learns to adapt the noise covariances of the filter itself, while EqNIO\cite{jayanth2025eqnio} improves generalization through rotation equivariance.
These methods target pedestrian or ground-robot motion, with limited vertical dynamics compared to unmanned aerial vehicles (UAVs).

Learning-based IO for UAVs adapts these ideas to the platform.
DIDO\cite{zhang2022dido} feeds the network the measured rotor speeds of a quadrotor, while IMO\cite{cioffi2023imo} feeds the collective thrust, both alongside the IMU.
DIVE\cite{bajwa2024dive} fuses a learned velocity into an inertial-only filter for quadrotors.
AirIO\cite{qiu2025airio} shows that keeping the IMU in its native body-frame representation lets the network capture the highly dynamic motion of drones.
AI-IO\cite{cui2026aiio} adds rotor-speed measurements to this body-frame formulation, motivated by quadrotor aerodynamics.
TartanIMU\cite{zhao2025tartanimu} generalizes across robotics platforms with a foundation model and platform-specific decoders.

These methods nonetheless frame velocity estimation as direct regression and take the measurement covariance from a separate decoder trained apart from the estimate.
We instead predict a probability distribution over discrete velocity bins, inspired by recent successes in monocular depth estimation (MDE)\cite{bhat2021adabins,li2024binsformer}.

\begin{figure*}[t!]
\begin{center}
\includegraphics[width=1\textwidth]{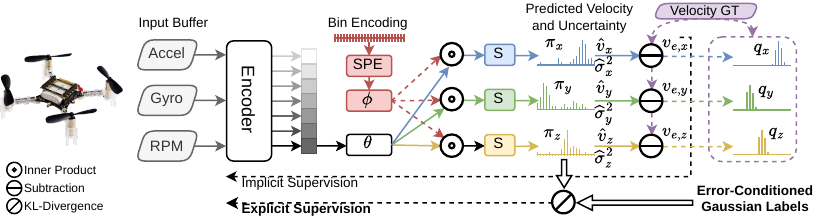}
\vspace{-18pt}
\caption{Overview of the VeloBins Architecture. A window of accelerometer, gyroscope, and rotor speed measurements is encoded into a per-axis bin distribution $\boldsymbol{\pi}_i$, whose logits pair the encoder query with a learnable bin encoding (Sec.~\ref{sec:bin_encoding}). The velocity $\hat{v}_i$ and its uncertainty $\hat{\sigma}^2_{i}$ are decoded from $\boldsymbol{\pi}_i$ (Sec.~\ref{sec:velocity_estimation_using_bins}). During training, the velocity error $v_{e,i}$ sets the width of the error-conditioned Gaussian label $\mathbf{q}_i$, which supervises $\boldsymbol{\pi}_i$ explicitly through a KL-divergence loss (Sec.~\ref{sec:bin_supervision}), while a Huber loss on $\hat{v}_i$ supervises it implicitly (Sec.~\ref{sec:training_objective}).}
\vspace{-22pt}
\label{fig:arch}
\end{center}
\end{figure*}

\subsection{Regression as a Hybrid Classification-Regression Task}

Bin-based methods reformulate a continuous target as classification over discretized bins\cite{zhang2023ordinal}.
DORN\cite{Fu_2018_dorn_depth} discretizes the depth range of MDE into ordered bins with an ordinal objective, while AdaBins\cite{bhat2021adabins} adapts the bin width per image and decodes the depth as an expectation over bin centers.
BinsFormer\cite{li2024binsformer} and IEBins\cite{shao2023iebins} refine the same design for MDE.
SimCC\cite{li2022simcc} and RTMPose\cite{jiang2023rtmpose} apply the same coordinate-classification view to human pose estimation.
Deep reinforcement learning models the value function as a categorical distribution\cite{bellemare2017distributional}, where classification outperforms regression at scale\cite{farebrother2024stopregressing}.

Another important aspect of bin-based methods is the supervision of the bin distribution.
Earlier works place no target on the distribution and supervise it through the decoded value, which leaves the variance unconstrained\cite{bhat2021adabins,li2024binsformer}.
SimCC\cite{li2022simcc} and RTMPose\cite{jiang2023rtmpose} supervise the distribution with a soft label of fixed width, which cannot reflect the per-sample uncertainty.
Distributional losses also set the target width heuristically, both in regression~\cite{imani2018histloss} and in reinforcement learning~\cite{farebrother2024stopregressing}.
RTMO\cite{lu2024rtmo} instead learns the target variance through a maximum-likelihood objective, which requires a separate decoder.
In this work, we explore the bin-based formulation for aerial inertial odometry, where an error-conditioned Gaussian label trains the bin variance to reflect the per-sample uncertainty without needing a separate decoder.

\section{Methodology}
As depicted in Fig.~\ref{fig:arch}, \textbf{VeloBins} reformulates velocity regression for inertial odometry as a hybrid classification-regression problem.
A single distribution over discretized velocity bins provides both the velocity and its uncertainty predictions.
An error-conditioned Gaussian label supervises the bin variance explicitly, so the uncertainty is learned rather than left unconstrained.
The EKF then consumes the two decoded quantities as its measurement and measurement covariance.

\subsection{Velocity and Uncertainty from Bins}
\label{sec:velocity_estimation_using_bins}

For each velocity axis $i \in \{x, y, z\}$, we define an operating range $[v_{\min}, v_{\max}]$ and discretize it into $N$ bins.
Let $\mathbf{b} = (b_0, b_1, \ldots, b_n, \ldots, b_{N-1})$ be the vector of bin centers.
Each bin $n$ covers a continuous range of velocities defined by its lower and upper boundaries, $[b^-_n,b^+_n]$.
These boundaries are set at the midpoint between adjacent centers, such that $b^+_n=\tfrac{1}{2}(b_n+b_{n+1})$ and $b^-_n=\tfrac{1}{2}(b_{n-1}+b_{n})$, with the boundaries of the first and last bins extending to $v_{\min}$ and $v_{\max}$, respectively.

The velocity decoder predicts a per-axis distribution $\boldsymbol{\pi}_{i}=\{\pi_{i,0},\dots,\pi_{i,N-1}\}$ over the bins and decodes the velocity as the expectation of this distribution over bin centers,
\begin{align}
    \hat{v}_{i \in \{x, y, z\}} = \sum_{n=0}^{N-1}\pi_{i,n} \, b_{n} .
\label{eq:decode}
\end{align}

Unlike prior learning-based IO methods that regress the uncertainty with a separate covariance decoder~\cite{liu2020tlio,qiu2025airio,cui2026aiio}, VeloBins decodes the velocity uncertainty from the same bin distribution, taking it as the variance about the decoded velocity $\hat{v}_i$,
\begin{align}
    \hat{\sigma}^2_{i \in \{x, y, z\}} = \sum_{n=0}^{N-1} \pi_{i,n} \, (b_{n} - \hat{v}_i)^2 .
\label{eq:bin_variance}
\end{align}

\subsection{VeloBins Network Architecture}
\label{sec:velobins_architecture}
The architecture of \textbf{VeloBins}, as shown in Fig.~\ref{fig:arch}, follows the encoder-decoder design of AI-IO~\cite{cui2026aiio} and replaces its velocity and covariance decoders with a single bin decoder.
Per-modality 1-D convolutions and a two-layer transformer encoder map the input window of accelerometer, gyroscope, and per-rotor actuation signals to a last-step feature $\mathbf{h} \in \mathbb{R}^{D_h}$.
We use the measured rotor speed where the platform provides ESC telemetry~\cite{cui2026aiio,bauersfeld2021neurobem}, and the commanded motor signal otherwise~\cite{ullah2026nanobench,bosello2024tii_ratm}.
A rotor follows its command through a first-order response and a platform-dependent map~\cite{bauersfeld2021neurobem}, which the encoder learns per dataset.

\subsubsection{Bin Encoding}
\label{sec:bin_encoding}
Rather than mapping $\mathbf{h}$ to $N$ logits with a linear layer~\cite{bhat2021adabins,li2024binsformer,jiang2023rtmpose}, we build each bin's classifier from an encoding of its coordinate~\cite{lu2024rtmo}, preserving the ordinal structure of the velocity range at a parameter count independent of $N$.
We encode each bin center with a learnable sine positional encoding (SPE)~\cite{sun2024spe}, scaling the fixed base by a learnable per-channel frequency $\boldsymbol{\gamma}$,
\begin{align}
    \mathrm{SPE}(\mathbf{b}) = \sin\!\left(\boldsymbol{\gamma}\mathrm{PE}(\mathbf{b})\right) \in \mathbb{R}^{N \times C},
\label{eq:spe}
\end{align}
where $\mathrm{PE}(\mathbf{b})$ is the fixed positional encoding of the bin centers $\mathbf{b}$ and $C$ is its channel dimension.
A learned linear map $\boldsymbol{\phi}$ then projects each bin's encoding to a lower dimension $D < C$, which forms the representation of that bin's classifier.

\subsubsection{Bin Classification}
For each velocity axis $i \in \{x, y, z\}$, a fully-connected layer $\boldsymbol{\theta}$ projects $\mathbf{h}$ into a query $\boldsymbol{\theta}_i(\mathbf{h}) \in \mathbb{R}^{D}$, which is matched against the projected encoding of each bin and normalized by a softmax~\cite{lu2024rtmo},
\begin{align}
    \pi_{i,n} = \frac{\exp\!\left(\boldsymbol{\theta}_i(\mathbf{h}) \cdot \boldsymbol{\phi}\!\left(\mathrm{SPE}(b_{n})\right)\right)}{\sum_{m=0}^{N-1} \exp\!\left(\boldsymbol{\theta}_i(\mathbf{h}) \cdot \boldsymbol{\phi}\!\left(\mathrm{SPE}(b_{m})\right)\right)}.
\label{eq:softmax}
\end{align}
The resulting $\boldsymbol{\pi}_i$ is then decoded into the velocity and its uncertainty by~(\ref{eq:decode}) and~(\ref{eq:bin_variance}).
Building each classifier from a bin coordinate also keeps the decoder small compared to an unconstrained linear layer, whose parameter count grows with the number of bins $N$.
With $D_h=48$, $C=64$, and $D=32$, it needs $C+D(C+1)+3D(D_h+1)=6.8$\,k parameters at any $N$, whereas the unconstrained layer needs $3N(D_h+1)=75.3$\,k at $N=512$.
The decoder therefore adds 9.8\% to the 69.8\,k parameters of the encoder.

\subsection{Error-Conditioned Gaussian Labels}
\label{sec:bin_supervision}
\begin{figure}[t]
    \centering
    \includegraphics[width=\columnwidth]{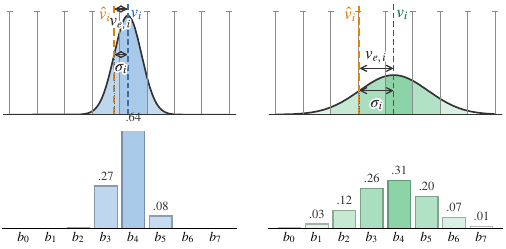}
    \vspace{-16pt}
    \caption{Visualization of error-conditioned Gaussian label construction. A smaller velocity error $v_{e,i}$ gives a sharp Gaussian label (left) and a larger one gives a label of larger standard deviation (right). The bottom panels show the resulting label $\mathbf{q}_i$ over the bins.}
    \label{fig:errcond}
    \vspace{-18pt}
\end{figure}

A regression loss such as L1, L2, or Huber on the decoded velocity $\hat{v}_i$ of~(\ref{eq:decode}) leaves the bin distribution variance unconstrained as many possible distributions share the same expected value.
To supervise the uncertainty explicitly, we construct an error-conditioned Gaussian $\bar{q}_i$ for each axis $i$, centered at the ground-truth velocity $v_i$ with a standard deviation $\sigma_i$ set by the velocity error $v_{e,i} = \left| \hat{v}_i - v_i \right|$ (Fig.~\ref{fig:errcond}),
\begin{align}
    \bar{q}_i = \mathcal{N}\!\left(v_i, \, \sigma_i^2\right), \qquad \sigma_i = v_{e,i},
\label{eq:errcond}
\end{align}
We detach $v_{e,i}$ when building the label, so no gradient flows through $\sigma_i$.
The velocity error $v_{e,i}$ is the per-sample stationary point of the Gaussian likelihood, and a zero-mean error gives $\mathbb{E}[\sigma_i^2] = \mathbb{E}[v_{e,i}^2] = \mathrm{Var}[\hat{v}_i - v_i]$, so the label variance is an unbiased target for the marginal error variance.
The EKF instead consumes the error variance conditioned on the input, so we verify per update whether the bin variance matches the error it incurs in Sec.~\ref{sec:calibration}.

The discrete label $\mathbf{q}_i$ integrates the continuous density $\bar{q}_i$ over each bin's support $[b^{-}_{n}, b^{+}_{n}]$,
\begin{align}
    q_{i,n} = \frac{1}{Z_i}\left[\;\Phi\!\left(\frac{b^{+}_{n} - v_i}{\sigma_i}\right) - \Phi\!\left(\frac{b^{-}_{n} - v_i}{\sigma_i}\right)\right],
\label{eq:target}
\end{align}
where $\Phi$ is the standard normal CDF and $Z_i$ normalizes the $N$ bins to sum to one.
Following the histogram loss of HL-Gauss~\cite{imani2018histloss,farebrother2024stopregressing}, each bin is assigned the probability \emph{mass} over its support rather than the density evaluated at its center, so adjacent bins retain non-zero mass and the ordinal structure of the velocity range is preserved.

However, the decoded mean of~(\ref{eq:target}) has a discretization bias~\cite{imani2026investigating} that grows as the label sharpens.
We therefore tilt $\mathbf{q}_i$ to the closest distribution in Kullback--Leibler (KL) divergence whose expectation is exactly $v_i$~\cite{csiszar1975idivergence},
\begin{align}
    q_{i,n} \;\leftarrow\; \frac{1}{\tilde{Z}_i} \, q_{i,n} \, \exp\!\left(\eta_i \, b_{n}\right),
\label{eq:tilt}
\end{align}
where the per-axis scalar $\eta_i$ is fixed by the mean constraint and solved outside the autograd graph, and $\tilde{Z}_i$ renormalizes the tilted label.
The correction changes the standard deviation $\sigma_i$ only to second order and vanishes ($\eta_i = 0$) when the decoded mean of~(\ref{eq:target}) is already exact.

\subsection{Training Objective}
\label{sec:training_objective}
The bin distribution is supervised by the KL divergence from the error-conditioned label to the prediction,
\begin{align}
    \mathcal{L}_{\mathrm{KL},i} = \mathrm{KL}\!\left(\mathbf{q}_i \, \| \, \boldsymbol{\pi}_i\right).
\label{eq:kl}
\end{align}
With $\sigma_i$ set by~(\ref{eq:errcond}), matching this label brings the bin variance of~(\ref{eq:bin_variance}) toward $\sigma_i^2$, so the velocity uncertainty is supervised explicitly rather than left as a by-product of training.
Since~(\ref{eq:kl}) is linear in the detached label $\mathbf{q}_i$, its per-input optimum is the conditional mean of the labels, thereby encouraging the predicted distribution to reflect the input-conditioned prediction error.

We also retain a Huber loss $\mathcal{L}_{\mathrm{H},i}$ on the decoded velocity $\hat{v}_i$~\cite{liu2020tlio,cui2026aiio}, which supervises $\boldsymbol{\pi}_i$ \emph{implicitly} through~(\ref{eq:decode}).
The KL divergence of~(\ref{eq:kl}) gives the decoded mean no special weight among the bins, whereas the Huber term with transition point $\delta$ penalizes a displaced mean directly,
\begin{align}
    \mathcal{L}_{\mathrm{H},i} =
    \begin{cases}
        \tfrac{1}{2} \, v_{e,i}^2, & v_{e,i} < \delta, \\[2pt]
        \delta \left( v_{e,i} - \tfrac{1}{2}\delta \right), & \text{otherwise}.
    \end{cases}
\label{eq:huber}
\end{align}
The total objective is
\begin{align}
    \mathcal{L} = \sum_{i \in \{x, y, z\}} \left( \lambda_{\mathrm{H}} \, \mathcal{L}_{\mathrm{H},i} + \lambda_{\mathrm{KL}} \, \mathcal{L}_{\mathrm{KL},i} \right),
\label{eq:total_loss}
\end{align}
where $\lambda_{\mathrm{H}}$ and $\lambda_{\mathrm{KL}}$ weight the Huber and KL terms.

\subsection{Extended Kalman Filter}
\label{sec:ekf}
We fuse the VeloBins prediction with the error-state EKF of AI-IO~\cite{cui2026aiio}, retaining its state definition, IMU propagation, and measurement-update formulation. The filter state is
$\mathbf{X}_k = (\mathbf{R}_k, \enspace {^G}\mathbf{v}_k, \enspace {^G}\mathbf{p}_k, \enspace \mathbf{b}_{a_k}, \enspace \mathbf{b}_{g_k})$, where \(\mathbf{R}_k\) is the body-to-global rotation, \({}^{G}\mathbf{v}_k\) and \({}^{G}\mathbf{p}_k\) are the global-frame velocity and position, and \(\mathbf{b}_{a,k}\) and \(\mathbf{b}_{g,k}\) are the IMU biases. We refer the reader to~\cite{cui2026aiio} for the complete propagation and EKF update equations.

At each network update, VeloBins provides the decoded body-frame velocity \(\hat{\mathbf{v}}_k\) of (1) as the measurement. Its observation model and covariance are
\begin{align}
    \mathbf{z}_k &= \hat{\mathbf{v}}_k, \enspace \enspace h(\mathbf{X}_k) = \mathbf{R}_k^T \cdot {^G}\mathbf{v}_k, \\ \hat{\boldsymbol{\Sigma}}_k &= \text{diag}\!\left(\hat{\sigma}^2_{x}, \enspace \hat{\sigma}^2_{y}, \enspace \hat{\sigma}^2_{z}\right),
\end{align}
where \(\hat{\sigma}_{i,k}^{2}\) is the bin variance decoded by (2). Thus, unlike AI-IO~\cite{cui2026aiio}, which predicts the velocity and covariance using separate regression decoders, VeloBins obtains both from the same bin distribution. The predicted covariance is directly used as the EKF measurement covariance without rescaling or innovation gating.

\section{Experiments}

\input{tables/datasets}
\subsection{Experimental Setup}

\input{tables/main_results}

\subsubsection{Datasets}
We evaluate on four public UAV datasets (Table~\ref{tab:datasets}), namely the AI-IO~\cite{cui2026aiio}, NanoBench~\cite{ullah2026nanobench}, TII-RATM~\cite{bosello2024tii_ratm}, and NeuroBEM~\cite{bauersfeld2021neurobem} datasets.
The platforms range from a 27~g Crazyflie to a 5-inch racing quadrotor, and supply either measured rotor speed or a commanded pulse-width modulation (PWM) signal.
The four test splits together contain 63 sequences of 5.3~km and 43~min of flight.

\subsubsection{Baselines}
We compare against five learning-based IO methods, all trained on the same data as VeloBins.
TLIO~\cite{liu2020tlio} and EqNIO~\cite{jayanth2025eqnio} are pedestrian displacement regressors, and for EqNIO we use the SO(2) equivariant-frame variant on the TLIO backbone.
IMO~\cite{cioffi2023imo} regresses displacement from gyroscope and collective thrust for drone racing, while AirIO~\cite{qiu2025airio} and AI-IO~\cite{cui2026aiio} predict body-frame velocity with a separate covariance decoder.

\subsubsection{Metrics}
We report the mean absolute velocity error (AVE, m/s) and the root mean square relative (RTE, m) and absolute (ATE, m) trajectory errors, with RTE evaluated over $\Delta t = 5$\,s.
We also report the negative log-likelihood (NLL)~\cite{loquercio2020uncertainty} to evaluate the predicted uncertainty, penalizing inaccurate estimates and over- or under-confident uncertainty together.
The median per-update velocity normalized estimation error squared (NEES)~\cite{barshalom2001estimation} then evaluates filter consistency after EKF fusion, whether the covariance the filter reports matches the error it incurs.

\subsubsection{Implementation Details}
We implement the network in PyTorch and train on an RTX 4090 with an i9-12900K CPU, using Adam at a constant learning rate of $3\times10^{-4}$ for 100 epochs at batch size 128.
We use $N=512$ bins per axis over a symmetric range $[-R, R]$ shared by all three axes, with $R \approx 1.1\times$ each training set's axis-wise maximum velocity.
The Huber transition point is $\delta = 0.1$\,m/s and $\lambda_{\mathrm{H}}=\lambda_{\mathrm{KL}}=1$ on every dataset.
We further clamp $\sigma_i$ to a minimum of one tenth of a bin width. 
\subsubsection{Inference and Runtime}
Following AI-IO~\cite{cui2026aiio}, the network takes a one-second window of 100~Hz inputs and supplies the EKF with velocity updates at 20~Hz.
On a single Cortex-A78AE CPU thread of an Orin NX computer, a forward pass takes 5.10~ms against 4.36~ms for the regression baseline, a 17\% overhead.
However, the bin encoding of~(\ref{eq:softmax}) are input-independent and can be precomputed once, which reduces the overhead to only 9\% (4.73~ms).

\begin{figure*}[t!]
\begin{center}
\includegraphics[width=\textwidth]{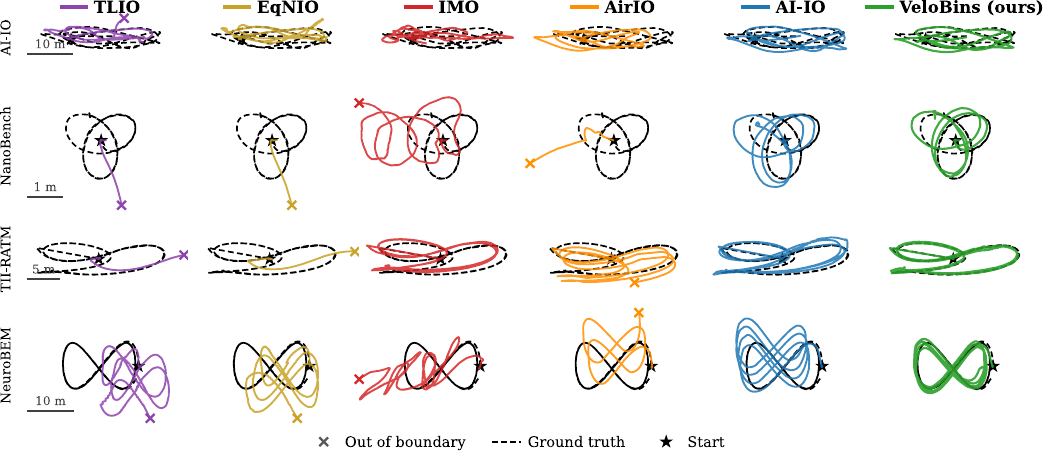}
\end{center}
\vspace{-8pt}
\caption{Qualitative comparison of VeloBins against the baselines. Each column shows one method's estimated trajectory on one held-out sequence per dataset, covering AI-IO, NanoBench, TII-RATM, and NeuroBEM datasets, against ground truth.}
\vspace{-4pt}
\label{fig:trajectories}
\end{figure*}

\input{tables/ablation_supervision}

\subsection{Inertial Odometry Accuracy}
\label{sec:results}

Table~\ref{tab:aggregated_result} reports EKF-fused results.
VeloBins achieves the lowest average AVE, RTE, and ATE on all four datasets.
Compared to the strongest baseline, AI-IO~\cite{cui2026aiio}, VeloBins lowers AVE by 3--27\%, RTE by 8--40\% and ATE by 6--53\% across datasets, with the largest gains on TII-RATM.
Since AI-IO and VeloBins share the same encoder, EKF modules, and training data, the improvements are attributable to the proposed bin reformulation and its error-conditioned Gaussian label supervision.

Against the remaining baselines, the margins are larger.
TLIO~\cite{liu2020tlio} and EqNIO~\cite{jayanth2025eqnio} are the least accurate on NanoBench, TII-RATM, and NeuroBEM, since their global-frame formulation does not capture the highly dynamic motion of aerial robots~\cite{qiu2025airio}.
IMO~\cite{cioffi2023imo} is competitive on NanoBench and TII-RATM thanks to its quadrotor-specific actuation inputs, but fails on the AI-IO dataset, where the free-form maneuvers reveal the overfitting of its data-intensive actuation-driven formulation.
AirIO~\cite{qiu2025airio} remains less accurate than AI-IO and VeloBins on every dataset, owing to its sensitivity to orientation errors and online windowed inference~\cite{cui2026aiio}.

Fig.~\ref{fig:trajectories} visualizes the estimated trajectories on one test sequence per dataset.
On these sequences, VeloBins drifts least and matches the ground truth most closely.

\section{Ablation Study}
\label{sec:ablation}
The ablations investigate the source of the improvements and measure how the supervision strategies and the bin-decoder design affect the velocity and uncertainty estimates, and in turn the fused trajectory errors and the filter consistency.
We examine four questions.
Does estimating velocity over bins yield a more accurate velocity estimate than direct regression (\textbf{Q1})?
How do the supervision strategies affect the network accuracy and the bin variance (\textbf{Q2})?
How does the predicted uncertainty affect the fused accuracy and the filter consistency (\textbf{Q3})?
How do the bin encoding and the bin resolution affect the estimate (\textbf{Q4})?

For \textbf{Q1}--\textbf{Q3}, Table~\ref{tab:ablation_obj} compares five bin-supervision variants against the regression baseline of AI-IO~\cite{cui2026aiio}.
They supervise the bin distribution with the Huber loss (\textbf{H}), the MLE loss of RTMO~\cite{lu2024rtmo} (\textbf{M}), or our error-conditioned Gaussian labels of Sec.~\ref{sec:bin_supervision} (\textbf{G}), the latter two also with the Huber term added (\textbf{+H}).
The superscript gives the uncertainty source used in fusion, either the bin variance ($^{B}$), the covariance decoder ($^{C}$), or their sum ($^{B+C}$).

\subsection{Regression vs.\ Classification}
To answer \textbf{Q1}, we compare the regression baseline against the bin-based formulation with Huber supervision (H).
Using bins with H lowers the network AVE on all four datasets, by 2\% on the AI-IO dataset up to 25\% on NanoBench.

\subsection{Supervision Strategies}
To answer \textbf{Q2}, we compare the network AVE and NLL across the supervision strategies in Table~\ref{tab:ablation_obj}.

The implicit supervision by \textbf{H} improves accuracy over the regression baseline, but its NLL is far worse than the regression baseline on every dataset, as it does not explicitly shape the uncertainty.
For the MLE supervision \textbf{M} and \textbf{M+H}, either source alone yields a higher NLL than the regression baseline, and only combining both ($\circ^{B+C}$) approaches it.

In contrast, the proposed \textbf{G}$^{B}$ and \textbf{G+H}$^{B}$ achieve a better-calibrated bin variance, with \textbf{G+H}$^{B}$ attaining the lowest NLL on all four datasets.
\textbf{Reg.}, \textbf{M}, and \textbf{M+H} all supervise their covariance with an NLL loss, yet they remain worse than the proposed variants that never optimize one.

The Huber term improves the proposed supervision, as \textbf{G+H} achieves a lower network AVE than \textbf{G} on all four datasets and also lowers the NLL on all four, whereas for \textbf{M} it helps neither consistently.

\begin{figure}[t]
\begin{center}
\includegraphics[width=\columnwidth]{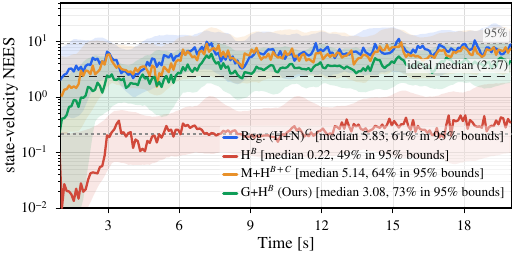}
\end{center}
\vspace{-8pt}
\caption{Per-update velocity NEES over the 63 test sequences across the datasets, showing the median and interquartile band against the ideal median and the 95\% bounds. The \textbf{VeloBins} with the proposed \textbf{G+H}$^{B}$ achieves the closest median NEES to the ideal and the most consistent filter updates with 73\% inside the 95\% bounds.}
\label{fig:nees}
\vspace{-4pt}
\end{figure}

\input{tables/ablation_encoding}

\subsection{Filter Accuracy and Consistency}
\label{sec:calibration}
To answer \textbf{Q3}, we evaluate the fused accuracy and the filter consistency of each variant.

After fusion, the network-AVE ordering reverses, as \textbf{G} and \textbf{G+H} attain the lowest fused AVE on three of the four datasets, with \textbf{G+H} attaining the lowest ATE on the AI-IO dataset and NanoBench.
This is attributable to the velocity uncertainty, since the EKF weights each update by it.
While \textbf{H} predicts an accurate velocity, its over-spread bin variance leads the filter to discard its updates.
The \textbf{M} and \textbf{M+H} variants are over-confident, like the regression baseline, while the proposed \textbf{G+H} is better-calibrated and ranks best overall.

Fig.~\ref{fig:nees} reports the per-update velocity NEES. The regression baseline and \textbf{M+H}$^{B+C}$ are over-confident and \textbf{H}$^{B}$ is under-confident.
\textbf{G+H}$^{B}$ is the closest to the ideal, with a median NEES of 3.08 against the ideal 2.37 and 73\% of updates inside the 95\% bounds, against 64\% for \textbf{M+H}$^{B+C}$, 61\% for the regression baseline, and 49\% for \textbf{H}$^{B}$.
The bin variance of \textbf{G+H}$^{B}$ therefore supplies the measurement covariance without a separate covariance decoder or an NLL loss.

\subsection{Bin-Decoder Design}
\label{sec:ablation_enc}
To answer \textbf{Q4}, we ablate the two structural choices of the bin decoder, which are the bin encoding and the bin resolution.

\subsubsection{Bin Encoding}
We vary how the logits are formed, comparing a \textbf{Direct} bin prediction~\cite{bhat2021adabins,li2022simcc}, the fixed positional encoding~\cite{lu2024rtmo} (\textbf{PE}), and learnable sine positional encoding \textbf{SPE} of~(\ref{eq:spe}).
As shown in Table~\ref{tab:ablation_enc}, \textbf{SPE} attains the lowest RTE on every dataset and the lowest NLL on three of the four. 
Note that, on TII-RATM, only \textbf{SPE} improves the NLL over the regression baseline~\cite{cui2026aiio}.
\textbf{Direct} discards the ordinal structure that a coordinate encoding preserves, while \textbf{PE} keeps the frequencies of its fixed base, and only the learnable $\boldsymbol{\gamma}$ of SPE~(\ref{eq:spe}) adapts the ordinal structures of the bins to the data~\cite{sun2024spe}.
The margin is largest on TII-RATM, where the highest speeds give the widest bin range and the error-conditioned label therefore spans the fewest bins.

\begin{figure}[t!]
\begin{center}
\includegraphics[width=\columnwidth]{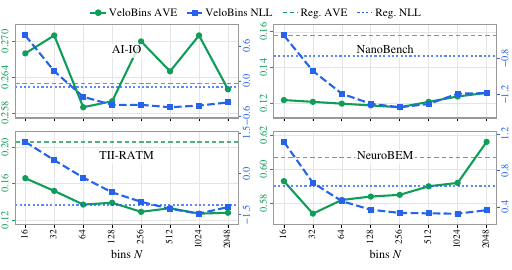}
\end{center}
\vspace{-12pt}
\caption{Bin-resolution ablation of the network AVE and NLL across varied bins $N$, per dataset, against the regression baseline. Accuracy varies little with $N$, while the NLL is better at $N>64$ and improves until $N\!=\!512$.}
\label{fig:bins}
\vspace{-14pt}
\end{figure}
\subsubsection{Bin Resolution}
\label{sec:ablation_bins}

Fig.~\ref{fig:bins} sweeps the bin number $N$ from 16 to 2048 against the regression baseline.
VeloBins achieves a lower network AVE at every $N$ on NanoBench and TII-RATM and stays within 0.01~m/s of the baseline on the other two datasets, so the velocity estimate does not need a fine grid.
However, the NLL consistently falls below the regression baseline for $N=64$ or above on three of the four datasets, and saturates by $N\!=\!512$, beyond which it changes little.
On TII-RATM the NLL stays higher than the regression baseline until $N\!=\!512$.
Once the velocity error becomes sufficiently small, the error-conditioned label collapses onto too few bins to supervise the bin variance reliably, leaving an under-confident uncertainty.
However, for EKF fusion, under-confidence is preferable to over-confidence, as over-confidence can lead to filter divergence~\cite{barshalom2001estimation}.

\section{Conclusion and Discussion}
\label{sec:discussion}

We presented VeloBins, which reformulates body-frame velocity estimation in inertial odometry as a distribution over discretized velocity bins.
The uncertainty is decoded from the same distribution, and supervised explicitly with error-conditioned Gaussian labels.
It achieves the lowest average velocity, relative, and absolute trajectory errors on four aerial datasets, reducing ATE by up to 53\% over the strongest baseline~\cite{cui2026aiio}.
We also demonstrated that VeloBins provides better-calibrated uncertainty than the evaluated alternatives, improving the consistency of downstream EKF fusion.

While our implementation and evaluation focused on aerial robots, we believe the formulation and its supervision are applicable to other platforms.
Humanoids~\cite{baumgartner2026cocoinekf}, quadrupeds~\cite{youm2025legged}, ground~\cite{brossard2020ai_imu_dr} and underwater vehicles~\cite{singh2025deepvl} each provide their own platform-specific modalities, or the method could be made cross-platform~\cite{zhao2025tartanimu}.

VeloBins has two limitations.
The bin grid is fixed per dataset with a known operational range, and once the velocity error becomes sufficiently small, the label collapses onto too few bins to supervise the variance reliably (Sec.~\ref{sec:ablation_bins}).








\bibliographystyle{ieeetr}
\bibliography{root}


\end{document}

%% file: tables/rankcolors.tex
\definecolor{r1_4}{rgb}{0.818,0.93,0.818}
\definecolor{r2_4}{rgb}{0.9393,0.9543,0.8273}
\definecolor{r3_4}{rgb}{0.9935,0.9309,0.8367}
\definecolor{r4_4}{rgb}{0.9804,0.86,0.846}

\definecolor{r1_5}{rgb}{0.818,0.93,0.818}
\definecolor{r2_5}{rgb}{0.909,0.9482,0.825}
\definecolor{r3_5}{rgb}{1.0,0.9664,0.832}
\definecolor{r4_5}{rgb}{0.9902,0.9132,0.839}
\definecolor{r5_5}{rgb}{0.9804,0.86,0.846}
\definecolor{a1_5}{rgb}{0.714,0.89,0.714}
\definecolor{a2_5}{rgb}{0.857,0.9186,0.725}
\definecolor{a3_5}{rgb}{1.0,0.9472,0.736}
\definecolor{a4_5}{rgb}{0.9846,0.8636,0.747}
\definecolor{a5_5}{rgb}{0.9692,0.78,0.758}

\definecolor{r1_6}{rgb}{0.818,0.93,0.818}
\definecolor{r2_6}{rgb}{0.8908,0.9446,0.8236}
\definecolor{r3_6}{rgb}{0.9636,0.9591,0.8292}
\definecolor{r4_6}{rgb}{0.9961,0.9451,0.8348}
\definecolor{r5_6}{rgb}{0.9882,0.9026,0.8404}
\definecolor{r6_6}{rgb}{0.9804,0.86,0.846}
\definecolor{a1_6}{rgb}{0.714,0.89,0.714}
\definecolor{a2_6}{rgb}{0.8284,0.9129,0.7228}
\definecolor{a3_6}{rgb}{0.9428,0.9358,0.7316}
\definecolor{a4_6}{rgb}{0.9938,0.9138,0.7404}
\definecolor{a5_6}{rgb}{0.9815,0.8469,0.7492}
\definecolor{a6_6}{rgb}{0.9692,0.78,0.758}

\definecolor{r1_9}{rgb}{0.818,0.93,0.818}
\definecolor{r2_9}{rgb}{0.8635,0.9391,0.8215}
\definecolor{r3_9}{rgb}{0.909,0.9482,0.825}
\definecolor{r4_9}{rgb}{0.9545,0.9573,0.8285}
\definecolor{r5_9}{rgb}{1.0,0.9664,0.832}
\definecolor{r6_9}{rgb}{0.9951,0.9398,0.8355}
\definecolor{r7_9}{rgb}{0.9902,0.9132,0.839}
\definecolor{r8_9}{rgb}{0.9853,0.8866,0.8425}
\definecolor{r9_9}{rgb}{0.9804,0.86,0.846}
\definecolor{a1_9}{rgb}{0.714,0.89,0.714}
\definecolor{a2_9}{rgb}{0.7855,0.9043,0.7195}
\definecolor{a3_9}{rgb}{0.857,0.9186,0.725}
\definecolor{a4_9}{rgb}{0.9285,0.9329,0.7305}
\definecolor{a5_9}{rgb}{1.0,0.9472,0.736}
\definecolor{a6_9}{rgb}{0.9923,0.9054,0.7415}
\definecolor{a7_9}{rgb}{0.9846,0.8636,0.747}
\definecolor{a8_9}{rgb}{0.9769,0.8218,0.7525}
\definecolor{a9_9}{rgb}{0.9692,0.78,0.758}

\definecolor{r1_10}{rgb}{0.818,0.93,0.818}
\definecolor{r2_10}{rgb}{0.8584,0.9381,0.8211}
\definecolor{r3_10}{rgb}{0.8989,0.9462,0.8242}
\definecolor{r4_10}{rgb}{0.9393,0.9543,0.8273}
\definecolor{r5_10}{rgb}{0.9798,0.9624,0.8304}
\definecolor{r6_10}{rgb}{0.9978,0.9546,0.8336}
\definecolor{r7_10}{rgb}{0.9935,0.9309,0.8367}
\definecolor{r8_10}{rgb}{0.9891,0.9073,0.8398}
\definecolor{r9_10}{rgb}{0.9848,0.8836,0.8429}
\definecolor{r10_10}{rgb}{0.9804,0.86,0.846}
\definecolor{a1_10}{rgb}{0.714,0.89,0.714}
\definecolor{a2_10}{rgb}{0.7776,0.9027,0.7189}
\definecolor{a3_10}{rgb}{0.8411,0.9154,0.7238}
\definecolor{a4_10}{rgb}{0.9047,0.9281,0.7287}
\definecolor{a5_10}{rgb}{0.9682,0.9408,0.7336}
\definecolor{a6_10}{rgb}{0.9966,0.9286,0.7384}
\definecolor{a7_10}{rgb}{0.9897,0.8915,0.7433}
\definecolor{a8_10}{rgb}{0.9829,0.8543,0.7482}
\definecolor{a9_10}{rgb}{0.976,0.8172,0.7531}
\definecolor{a10_10}{rgb}{0.9692,0.78,0.758}

%% file: tables/datasets.tex
\begin{table}[t!]
\centering
\caption{Dataset details. Len and Dur are the trajectory length and duration of the held-out test sequences.}
\label{tab:datasets}
\vspace{-4pt}
\footnotesize
\setlength{\tabcolsep}{1.5pt}
\renewcommand{\arraystretch}{1.15}
\begin{tabular*}{\columnwidth}{@{\extracolsep{\fill}}l c ccc cc cc cc}
\hline
\multirow{2}{*}{Dataset} & Rotor & \multicolumn{3}{c}{Sequences} & \multicolumn{2}{c}{$v$ [m/s]} & \multicolumn{2}{c}{Len [m]} & \multicolumn{2}{c}{Dur [s]} \\
\cline{3-5}\cline{6-7}\cline{8-9}\cline{10-11}
 & input & Train & Val & Test & $\bar{v}$ & $v_{\max}$ & mean & total & mean & total \\
\hline
AI-IO~\cite{cui2026aiio}                & RPM    &  12 & 12 & 22 & 1.5 & 13.6 &  65 & 1430 & 44 &  969 \\
NanoBench~\cite{ullah2026nanobench}     & PWM    &  81 &  4 & 22 & 0.5 &  2.0 &  19 &  411 & 41 &  909 \\
TII-RATM~\cite{bosello2024tii_ratm}     & PWM    &   9 &  3 &  6 & 3.5 & 21.8 & 110 &  661 & 31 &  188 \\
NeuroBEM~\cite{bauersfeld2021neurobem}  & RPM    &  47 & 16 & 13 & 5.1 & 17.7 & 215 & 2791 & 41 &  539 \\
\hline
\end{tabular*}
\vspace{-14pt}
\end{table}

%% file: tables/main_results.tex
\begin{table*}[t!]
    \centering
    \caption{Inertial Odometry AVE [m/s], RTE [m], and ATE [m] across the four datasets. Cells are shaded by \emph{rank} across the six methods within each row and metric (green\,=\,\textbf{best} $\rightarrow$ red\,=\,worst). All results are EKF-fused.}
    \label{tab:aggregated_result}
    \renewcommand{\arraystretch}{1.15}
    \setlength{\tabcolsep}{2.2pt}
    \footnotesize
    \begin{tabular*}{\textwidth}{@{\extracolsep{\fill}}c l | ccc | ccc | ccc | ccc | ccc | ccc}
    \hline\hline
    \multicolumn{2}{c|}{\multirow{2}{*}{Dataset / Shape}} & \multicolumn{3}{c|}{TLIO} & \multicolumn{3}{c|}{EqNIO} & \multicolumn{3}{c|}{IMO} & \multicolumn{3}{c|}{AirIO} & \multicolumn{3}{c|}{AI-IO} & \multicolumn{3}{c}{VeloBins} \\
    \cline{3-5}\cline{6-8}\cline{9-11}\cline{12-14}\cline{15-17}\cline{18-20}
    \multicolumn{2}{c|}{} & AVE & RTE & ATE & AVE & RTE & ATE & AVE & RTE & ATE & AVE & RTE & ATE & AVE & RTE & ATE & AVE & RTE & ATE \\
    \hline\hline
    \multirow{9}{*}{\rotatebox[origin=c]{90}{AI-IO}} & circle $\times$2 & \cellcolor{r5_6}0.898 & \cellcolor{r5_6}4.278 & \cellcolor{r5_6}7.673 & \cellcolor{r4_6}0.790 & \cellcolor{r4_6}3.973 & \cellcolor{r3_6}5.315 & \cellcolor{r6_6}5.509 & \cellcolor{r6_6}8.155 & \cellcolor{r6_6}13.056 & \cellcolor{r3_6}0.320 & \cellcolor{r3_6}1.278 & \cellcolor{r4_6}6.685 & \cellcolor{r1_6}\textbf{0.162} & \cellcolor{r1_6}\textbf{0.653} & \cellcolor{r2_6}2.456 & \cellcolor{r2_6}0.163 & \cellcolor{r2_6}0.669 & \cellcolor{r1_6}\textbf{2.111} \\
     & eight $\times$2 & \cellcolor{r5_6}1.005 & \cellcolor{r4_6}3.512 & \cellcolor{r3_6}8.702 & \cellcolor{r4_6}0.993 & \cellcolor{r5_6}3.668 & \cellcolor{r4_6}9.056 & \cellcolor{r6_6}5.034 & \cellcolor{r6_6}6.900 & \cellcolor{r6_6}17.926 & \cellcolor{r3_6}0.469 & \cellcolor{r3_6}2.022 & \cellcolor{r5_6}13.111 & \cellcolor{r2_6}0.240 & \cellcolor{r2_6}1.034 & \cellcolor{r2_6}6.284 & \cellcolor{r1_6}\textbf{0.204} & \cellcolor{r1_6}\textbf{0.786} & \cellcolor{r1_6}\textbf{4.723} \\
     & man/high $\times$4 & \cellcolor{r5_6}0.888 & \cellcolor{r5_6}4.349 & \cellcolor{r5_6}5.669 & \cellcolor{r3_6}0.706 & \cellcolor{r4_6}3.427 & \cellcolor{r3_6}3.710 & \cellcolor{r6_6}4.732 & \cellcolor{r6_6}8.307 & \cellcolor{r6_6}7.139 & \cellcolor{r4_6}0.854 & \cellcolor{r3_6}2.324 & \cellcolor{r4_6}3.984 & \cellcolor{r1_6}\textbf{0.340} & \cellcolor{r2_6}1.106 & \cellcolor{r2_6}1.684 & \cellcolor{r2_6}0.342 & \cellcolor{r1_6}\textbf{1.032} & \cellcolor{r1_6}\textbf{1.270} \\
     & man/low $\times$4 & \cellcolor{r5_6}0.837 & \cellcolor{r5_6}4.137 & \cellcolor{r5_6}4.949 & \cellcolor{r4_6}0.733 & \cellcolor{r4_6}3.813 & \cellcolor{r4_6}3.861 & \cellcolor{r6_6}4.470 & \cellcolor{r6_6}7.264 & \cellcolor{r6_6}8.909 & \cellcolor{r3_6}0.552 & \cellcolor{r3_6}1.424 & \cellcolor{r3_6}2.605 & \cellcolor{r2_6}0.243 & \cellcolor{r2_6}0.718 & \cellcolor{r2_6}1.365 & \cellcolor{r1_6}\textbf{0.232} & \cellcolor{r1_6}\textbf{0.657} & \cellcolor{r1_6}\textbf{1.272} \\
     & man/med $\times$4 & \cellcolor{r5_6}0.983 & \cellcolor{r5_6}4.030 & \cellcolor{r5_6}5.056 & \cellcolor{r4_6}0.862 & \cellcolor{r4_6}3.223 & \cellcolor{r4_6}3.749 & \cellcolor{r6_6}6.049 & \cellcolor{r6_6}11.833 & \cellcolor{r6_6}12.423 & \cellcolor{r3_6}0.566 & \cellcolor{r3_6}1.466 & \cellcolor{r3_6}2.584 & \cellcolor{r2_6}0.263 & \cellcolor{r2_6}0.758 & \cellcolor{r2_6}1.379 & \cellcolor{r1_6}\textbf{0.258} & \cellcolor{r1_6}\textbf{0.677} & \cellcolor{r1_6}\textbf{1.131} \\
     & random $\times$2 & \cellcolor{r5_6}0.882 & \cellcolor{r5_6}2.739 & \cellcolor{r4_6}6.098 & \cellcolor{r4_6}0.756 & \cellcolor{r4_6}2.474 & \cellcolor{r5_6}7.560 & \cellcolor{r6_6}4.476 & \cellcolor{r6_6}6.215 & \cellcolor{r6_6}13.528 & \cellcolor{r3_6}0.361 & \cellcolor{r3_6}1.030 & \cellcolor{r3_6}5.173 & \cellcolor{r1_6}\textbf{0.193} & \cellcolor{r1_6}\textbf{0.727} & \cellcolor{r1_6}\textbf{3.979} & \cellcolor{r2_6}0.205 & \cellcolor{r2_6}0.815 & \cellcolor{r2_6}4.710 \\
     & ud/circ $\times$2 & \cellcolor{r5_6}1.053 & \cellcolor{r5_6}4.475 & \cellcolor{r5_6}10.831 & \cellcolor{r4_6}0.983 & \cellcolor{r4_6}4.061 & \cellcolor{r3_6}6.619 & \cellcolor{r6_6}5.193 & \cellcolor{r6_6}6.625 & \cellcolor{r6_6}11.992 & \cellcolor{r3_6}0.375 & \cellcolor{r3_6}1.478 & \cellcolor{r4_6}8.219 & \cellcolor{r2_6}0.201 & \cellcolor{r2_6}0.839 & \cellcolor{r2_6}4.766 & \cellcolor{r1_6}\textbf{0.191} & \cellcolor{r1_6}\textbf{0.762} & \cellcolor{r1_6}\textbf{4.094} \\
     & ud/eight $\times$2 & \cellcolor{r5_6}0.815 & \cellcolor{r4_6}2.954 & \cellcolor{r3_6}6.398 & \cellcolor{r4_6}0.765 & \cellcolor{r5_6}3.231 & \cellcolor{r5_6}6.920 & \cellcolor{r6_6}4.366 & \cellcolor{r6_6}8.496 & \cellcolor{r6_6}23.319 & \cellcolor{r3_6}0.314 & \cellcolor{r3_6}1.167 & \cellcolor{r4_6}6.753 & \cellcolor{r2_6}0.186 & \cellcolor{r2_6}0.613 & \cellcolor{r2_6}2.930 & \cellcolor{r1_6}\textbf{0.161} & \cellcolor{r1_6}\textbf{0.485} & \cellcolor{r1_6}\textbf{1.631} \\
    \cline{2-20}
     & \textbf{AVG} & \cellcolor{a5_6}0.916 & \cellcolor{a5_6}3.908 & \cellcolor{a5_6}6.459 & \cellcolor{a4_6}0.808 & \cellcolor{a4_6}3.485 & \cellcolor{a3_6}5.283 & \cellcolor{a6_6}5.007 & \cellcolor{a6_6}8.291 & \cellcolor{a6_6}12.433 & \cellcolor{a3_6}0.526 & \cellcolor{a3_6}1.582 & \cellcolor{a4_6}5.299 & \cellcolor{a2_6}0.243 & \cellcolor{a2_6}0.821 & \cellcolor{a2_6}2.661 & \cellcolor{a1_6}\textbf{0.235} & \cellcolor{a1_6}\textbf{0.750} & \cellcolor{a1_6}\textbf{2.238} \\
    \hline\hline
    \multirow{8}{*}{\rotatebox[origin=c]{90}{NanoBench}} & circle $\times$3 & \cellcolor{r6_6}3.280 & \cellcolor{r6_6}8.782 & \cellcolor{r6_6}9.421 & \cellcolor{r5_6}2.604 & \cellcolor{r5_6}5.272 & \cellcolor{r4_6}5.611 & \cellcolor{r3_6}0.649 & \cellcolor{r3_6}0.934 & \cellcolor{r3_6}1.431 & \cellcolor{r4_6}0.785 & \cellcolor{r4_6}3.708 & \cellcolor{r5_6}7.201 & \cellcolor{r2_6}0.246 & \cellcolor{r1_6}\textbf{0.596} & \cellcolor{r1_6}\textbf{0.598} & \cellcolor{r1_6}\textbf{0.232} & \cellcolor{r2_6}0.613 & \cellcolor{r2_6}0.661 \\
     & figure8 $\times$3 & \cellcolor{r6_6}1.617 & \cellcolor{r6_6}8.035 & \cellcolor{r6_6}8.394 & \cellcolor{r5_6}1.541 & \cellcolor{r5_6}6.769 & \cellcolor{r5_6}6.166 & \cellcolor{r3_6}0.441 & \cellcolor{r3_6}0.828 & \cellcolor{r3_6}1.143 & \cellcolor{r4_6}0.590 & \cellcolor{r4_6}2.190 & \cellcolor{r4_6}3.332 & \cellcolor{r2_6}0.147 & \cellcolor{r2_6}0.469 & \cellcolor{r2_6}0.631 & \cellcolor{r1_6}\textbf{0.128} & \cellcolor{r1_6}\textbf{0.382} & \cellcolor{r1_6}\textbf{0.562} \\
     & helix $\times$3 & \cellcolor{r6_6}0.653 & \cellcolor{r5_6}1.684 & \cellcolor{r5_6}2.293 & \cellcolor{r5_6}0.608 & \cellcolor{r4_6}1.555 & \cellcolor{r3_6}1.902 & \cellcolor{r3_6}0.357 & \cellcolor{r3_6}0.905 & \cellcolor{r4_6}2.045 & \cellcolor{r4_6}0.492 & \cellcolor{r6_6}2.168 & \cellcolor{r6_6}7.055 & \cellcolor{r2_6}0.145 & \cellcolor{r2_6}0.532 & \cellcolor{r2_6}1.636 & \cellcolor{r1_6}\textbf{0.112} & \cellcolor{r1_6}\textbf{0.401} & \cellcolor{r1_6}\textbf{1.179} \\
     & oval $\times$3 & \cellcolor{r6_6}1.272 & \cellcolor{r6_6}2.998 & \cellcolor{r5_6}3.246 & \cellcolor{r5_6}1.220 & \cellcolor{r5_6}2.942 & \cellcolor{r4_6}2.574 & \cellcolor{r3_6}0.414 & \cellcolor{r3_6}0.473 & \cellcolor{r3_6}1.132 & \cellcolor{r4_6}0.524 & \cellcolor{r4_6}1.871 & \cellcolor{r6_6}6.295 & \cellcolor{r2_6}0.112 & \cellcolor{r2_6}0.215 & \cellcolor{r1_6}\textbf{0.327} & \cellcolor{r1_6}\textbf{0.092} & \cellcolor{r1_6}\textbf{0.159} & \cellcolor{r2_6}0.351 \\
     & star $\times$3 & \cellcolor{r6_6}2.902 & \cellcolor{r5_6}5.954 & \cellcolor{r5_6}10.765 & \cellcolor{r5_6}2.814 & \cellcolor{r6_6}6.041 & \cellcolor{r4_6}8.379 & \cellcolor{r3_6}0.473 & \cellcolor{r3_6}0.579 & \cellcolor{r3_6}1.274 & \cellcolor{r4_6}0.809 & \cellcolor{r4_6}3.971 & \cellcolor{r6_6}11.194 & \cellcolor{r2_6}0.150 & \cellcolor{r2_6}0.326 & \cellcolor{r2_6}0.824 & \cellcolor{r1_6}\textbf{0.115} & \cellcolor{r1_6}\textbf{0.279} & \cellcolor{r1_6}\textbf{0.823} \\
     & trefoil $\times$3 & \cellcolor{r6_6}1.084 & \cellcolor{r6_6}2.861 & \cellcolor{r5_6}2.860 & \cellcolor{r5_6}0.921 & \cellcolor{r4_6}2.358 & \cellcolor{r4_6}2.260 & \cellcolor{r3_6}0.329 & \cellcolor{r3_6}0.715 & \cellcolor{r2_6}1.374 & \cellcolor{r4_6}0.614 & \cellcolor{r5_6}2.810 & \cellcolor{r6_6}8.649 & \cellcolor{r2_6}0.161 & \cellcolor{r2_6}0.600 & \cellcolor{r3_6}1.478 & \cellcolor{r1_6}\textbf{0.117} & \cellcolor{r1_6}\textbf{0.412} & \cellcolor{r1_6}\textbf{0.923} \\
     & other $\times$4 & \cellcolor{r6_6}5.647 & \cellcolor{r6_6}36.042 & \cellcolor{r6_6}48.353 & \cellcolor{r5_6}3.910 & \cellcolor{r5_6}33.080 & \cellcolor{r5_6}33.860 & \cellcolor{r4_6}1.768 & \cellcolor{r3_6}1.609 & \cellcolor{r3_6}3.973 & \cellcolor{r3_6}0.662 & \cellcolor{r4_6}3.107 & \cellcolor{r4_6}19.604 & \cellcolor{r2_6}0.356 & \cellcolor{r2_6}0.907 & \cellcolor{r2_6}2.716 & \cellcolor{r1_6}\textbf{0.321} & \cellcolor{r1_6}\textbf{0.734} & \cellcolor{r1_6}\textbf{1.972} \\
    \cline{2-20}
     & \textbf{AVG} & \cellcolor{a6_6}2.501 & \cellcolor{a6_6}10.687 & \cellcolor{a6_6}13.834 & \cellcolor{a5_6}2.035 & \cellcolor{a5_6}9.415 & \cellcolor{a5_6}9.823 & \cellcolor{a4_6}0.685 & \cellcolor{a3_6}0.897 & \cellcolor{a3_6}1.868 & \cellcolor{a3_6}0.640 & \cellcolor{a4_6}2.845 & \cellcolor{a4_6}9.527 & \cellcolor{a2_6}0.196 & \cellcolor{a2_6}0.538 & \cellcolor{a2_6}1.243 & \cellcolor{a1_6}\textbf{0.167} & \cellcolor{a1_6}\textbf{0.440} & \cellcolor{a1_6}\textbf{0.972} \\
    \hline\hline
    \multirow{4}{*}{\rotatebox[origin=c]{90}{TII-RATM}} & ellipse $\times$2 & \cellcolor{r6_6}1.012 & \cellcolor{r6_6}4.888 & \cellcolor{r6_6}4.122 & \cellcolor{r5_6}0.944 & \cellcolor{r5_6}4.591 & \cellcolor{r5_6}3.606 & \cellcolor{r3_6}0.422 & \cellcolor{r3_6}0.832 & \cellcolor{r3_6}0.899 & \cellcolor{r4_6}0.775 & \cellcolor{r4_6}1.980 & \cellcolor{r4_6}3.259 & \cellcolor{r2_6}0.153 & \cellcolor{r2_6}0.338 & \cellcolor{r2_6}0.863 & \cellcolor{r1_6}\textbf{0.119} & \cellcolor{r1_6}\textbf{0.214} & \cellcolor{r1_6}\textbf{0.335} \\
     & lemniscate $\times$2 & \cellcolor{r6_6}0.916 & \cellcolor{r6_6}5.017 & \cellcolor{r6_6}3.261 & \cellcolor{r5_6}0.730 & \cellcolor{r5_6}4.475 & \cellcolor{r5_6}3.102 & \cellcolor{r3_6}0.334 & \cellcolor{r3_6}0.516 & \cellcolor{r3_6}0.781 & \cellcolor{r4_6}0.358 & \cellcolor{r4_6}0.630 & \cellcolor{r4_6}1.007 & \cellcolor{r2_6}0.130 & \cellcolor{r2_6}0.219 & \cellcolor{r2_6}0.395 & \cellcolor{r1_6}\textbf{0.095} & \cellcolor{r1_6}\textbf{0.173} & \cellcolor{r1_6}\textbf{0.269} \\
     & trackRATM $\times$2 & \cellcolor{r5_6}1.539 & \cellcolor{r6_6}9.282 & \cellcolor{r6_6}12.963 & \cellcolor{r6_6}1.707 & \cellcolor{r5_6}9.246 & \cellcolor{r5_6}8.387 & \cellcolor{r3_6}0.478 & \cellcolor{r3_6}0.951 & \cellcolor{r3_6}2.390 & \cellcolor{r4_6}0.642 & \cellcolor{r4_6}1.948 & \cellcolor{r4_6}4.788 & \cellcolor{r2_6}0.189 & \cellcolor{r2_6}0.468 & \cellcolor{r2_6}1.212 & \cellcolor{r1_6}\textbf{0.127} & \cellcolor{r1_6}\textbf{0.229} & \cellcolor{r1_6}\textbf{0.553} \\
    \cline{2-20}
     & \textbf{AVG} & \cellcolor{a6_6}1.156 & \cellcolor{a6_6}6.396 & \cellcolor{a6_6}6.782 & \cellcolor{a5_6}1.127 & \cellcolor{a5_6}6.104 & \cellcolor{a5_6}5.032 & \cellcolor{a3_6}0.411 & \cellcolor{a3_6}0.766 & \cellcolor{a3_6}1.357 & \cellcolor{a4_6}0.592 & \cellcolor{a4_6}1.519 & \cellcolor{a4_6}3.018 & \cellcolor{a2_6}0.157 & \cellcolor{a2_6}0.342 & \cellcolor{a2_6}0.823 & \cellcolor{a1_6}\textbf{0.114} & \cellcolor{a1_6}\textbf{0.206} & \cellcolor{a1_6}\textbf{0.386} \\
    \hline\hline
    \multirow{6}{*}{\rotatebox[origin=c]{90}{NeuroBEM}} & 3D circle $\times$2 & \cellcolor{r6_6}2.422 & \cellcolor{r5_6}3.328 & \cellcolor{r5_6}6.124 & \cellcolor{r5_6}2.335 & \cellcolor{r4_6}3.066 & \cellcolor{r2_6}3.808 & \cellcolor{r4_6}1.294 & \cellcolor{r6_6}3.478 & \cellcolor{r6_6}7.767 & \cellcolor{r3_6}0.616 & \cellcolor{r3_6}2.103 & \cellcolor{r3_6}4.787 & \cellcolor{r2_6}0.443 & \cellcolor{r2_6}1.040 & \cellcolor{r4_6}4.870 & \cellcolor{r1_6}\textbf{0.378} & \cellcolor{r1_6}\textbf{0.815} & \cellcolor{r1_6}\textbf{3.542} \\
     & lemniscate $\times$4 & \cellcolor{r4_6}1.621 & \cellcolor{r6_6}6.057 & \cellcolor{r6_6}9.841 & \cellcolor{r5_6}1.704 & \cellcolor{r5_6}5.361 & \cellcolor{r3_6}6.468 & \cellcolor{r6_6}2.153 & \cellcolor{r4_6}4.964 & \cellcolor{r4_6}7.113 & \cellcolor{r3_6}1.077 & \cellcolor{r3_6}3.132 & \cellcolor{r5_6}8.876 & \cellcolor{r2_6}0.585 & \cellcolor{r2_6}1.124 & \cellcolor{r2_6}4.198 & \cellcolor{r1_6}\textbf{0.496} & \cellcolor{r1_6}\textbf{0.889} & \cellcolor{r1_6}\textbf{3.081} \\
     & pow\_climb $\times$2 & \cellcolor{r5_6}2.393 & \cellcolor{r6_6}9.029 & \cellcolor{r6_6}10.522 & \cellcolor{r6_6}2.448 & \cellcolor{r5_6}6.340 & \cellcolor{r5_6}8.732 & \cellcolor{r3_6}1.091 & \cellcolor{r3_6}2.353 & \cellcolor{r1_6}\textbf{2.240} & \cellcolor{r4_6}1.148 & \cellcolor{r4_6}3.193 & \cellcolor{r4_6}8.205 & \cellcolor{r1_6}\textbf{0.542} & \cellcolor{r1_6}\textbf{0.807} & \cellcolor{r2_6}2.784 & \cellcolor{r2_6}0.544 & \cellcolor{r2_6}0.983 & \cellcolor{r3_6}3.843 \\
     & satellite $\times$2 & \cellcolor{r6_6}5.865 & \cellcolor{r6_6}8.696 & \cellcolor{r6_6}23.419 & \cellcolor{r5_6}4.739 & \cellcolor{r5_6}7.323 & \cellcolor{r4_6}8.887 & \cellcolor{r4_6}1.172 & \cellcolor{r3_6}2.306 & \cellcolor{r3_6}3.571 & \cellcolor{r3_6}1.024 & \cellcolor{r4_6}4.140 & \cellcolor{r5_6}9.155 & \cellcolor{r1_6}\textbf{0.511} & \cellcolor{r1_6}\textbf{0.968} & \cellcolor{r1_6}\textbf{2.811} & \cellcolor{r2_6}0.514 & \cellcolor{r2_6}0.977 & \cellcolor{r2_6}2.944 \\
     & other $\times$3 & \cellcolor{r5_6}1.297 & \cellcolor{r6_6}5.901 & \cellcolor{r4_6}11.227 & \cellcolor{r3_6}1.291 & \cellcolor{r5_6}5.524 & \cellcolor{r3_6}11.000 & \cellcolor{r6_6}2.065 & \cellcolor{r4_6}5.289 & \cellcolor{r6_6}13.148 & \cellcolor{r4_6}1.293 & \cellcolor{r3_6}3.097 & \cellcolor{r5_6}12.140 & \cellcolor{r1_6}\textbf{0.719} & \cellcolor{r2_6}1.134 & \cellcolor{r1_6}\textbf{7.357} & \cellcolor{r2_6}0.726 & \cellcolor{r1_6}\textbf{1.117} & \cellcolor{r2_6}7.736 \\
    \cline{2-20}
     & \textbf{AVG} & \cellcolor{a6_6}2.441 & \cellcolor{a6_6}6.465 & \cellcolor{a6_6}11.783 & \cellcolor{a5_6}2.287 & \cellcolor{a5_6}5.498 & \cellcolor{a4_6}7.825 & \cellcolor{a4_6}1.686 & \cellcolor{a4_6}4.000 & \cellcolor{a3_6}7.312 & \cellcolor{a3_6}1.059 & \cellcolor{a3_6}3.130 & \cellcolor{a5_6}8.940 & \cellcolor{a2_6}0.576 & \cellcolor{a2_6}1.041 & \cellcolor{a2_6}4.599 & \cellcolor{a1_6}\textbf{0.541} & \cellcolor{a1_6}\textbf{0.958} & \cellcolor{a1_6}\textbf{4.322} \\
    \hline\hline
    \hline
    \end{tabular*}
    \vspace{-8pt}
\end{table*}

%% file: tables/ablation_supervision.tex
\begin{table*}[t!]
    \centering
    \caption{Supervision strategy ablation. \textbf{Net} and \textbf{EKF} columns are the raw network and the fused outputs, respectively. \emph{vs Reg.} counts the metrics beating the regression baseline out of 20, and \emph{rank} is the average rank, lower being better. Cells are shaded by rank within each column, best in bold. The proposed \textbf{VeloBins} with \textbf{G+H}$^{B}$ ranks best overall.}
    \label{tab:ablation_obj}
    \vspace{-6pt}
    \renewcommand{\arraystretch}{1.15}
    \setlength{\tabcolsep}{0.9pt}
    \scriptsize
    \begin{tabular*}{\textwidth}{@{\extracolsep{\fill}}c l | ccccc | ccccc | ccccc | ccccc | cc}
    \hline\hline
    & \multirow{3}{*}{Loss$^{\dagger}$} & \multicolumn{5}{c|}{AI-IO} & \multicolumn{5}{c|}{NanoBench} & \multicolumn{5}{c|}{TII-RATM} & \multicolumn{5}{c|}{NeuroBEM} & \multicolumn{2}{c}{Overall} \\
    \cline{3-7}\cline{8-12}\cline{13-17}\cline{18-22}\cline{23-24}
     & & \multicolumn{2}{c}{Net} & \multicolumn{3}{c|}{EKF} & \multicolumn{2}{c}{Net} & \multicolumn{3}{c|}{EKF} & \multicolumn{2}{c}{Net} & \multicolumn{3}{c|}{EKF} & \multicolumn{2}{c}{Net} & \multicolumn{3}{c|}{EKF} & \multirow{2}{*}{vs Reg.} & \multirow{2}{*}{rank} \\
    \cline{3-4}\cline{5-7}\cline{8-9}\cline{10-12}\cline{13-14}\cline{15-17}\cline{18-19}\cline{20-22}
     & & AVE & NLL & AVE & RTE & ATE & AVE & NLL & AVE & RTE & ATE & AVE & NLL & AVE & RTE & ATE & AVE & NLL & AVE & RTE & ATE & & \\
    \hline\hline
     \multicolumn{2}{c|}{Reg. (H+N)$^{C}$} & \cellcolor{r5_10}0.263 & \cellcolor{r4_10}-0.11 & \cellcolor{r5_10}0.243 & \cellcolor{r6_10}0.821 & \cellcolor{r6_10}2.661 & \cellcolor{r10_10}0.158 & \cellcolor{r4_10}-0.78 & \cellcolor{r5_10}0.196 & \cellcolor{r10_10}0.539 & \cellcolor{r10_10}1.243 & \cellcolor{r10_10}0.206 & \cellcolor{r4_10}-1.15 & \cellcolor{r9_10}0.157 & \cellcolor{r9_10}0.342 & \cellcolor{r9_10}0.823 & \cellcolor{r9_10}0.607 & \cellcolor{r5_10}0.63 & \cellcolor{r6_10}0.576 & \cellcolor{r9_10}1.041 & \cellcolor{r9_10}4.599 &  & \cellcolor{r9_10}7.2 \\
    \hline
    \multirow{9}{*}{\rotatebox[origin=c]{90}{VeloBins}} & H$^{B}$ & \cellcolor{r1_10}\textbf{0.257} & \cellcolor{r8_10}2.62 & \cellcolor{r10_10}0.336 & \cellcolor{r9_10}0.865 & \cellcolor{r7_10}2.684 & \cellcolor{r1_10}\textbf{0.118} & \cellcolor{r8_10}0.30 & \cellcolor{r10_10}0.293 & \cellcolor{r9_10}0.517 & \cellcolor{r7_10}1.101 & \cellcolor{r6_10}0.173 & \cellcolor{r9_10}2.05 & \cellcolor{r10_10}0.532 & \cellcolor{r10_10}1.944 & \cellcolor{r10_10}5.211 & \cellcolor{r1_10}\textbf{0.570} & \cellcolor{r6_10}2.91 & \cellcolor{r10_10}0.763 & \cellcolor{r10_10}1.302 & \cellcolor{r10_10}6.242 & \cellcolor{r9_9}6/20 & \cellcolor{r10_10}7.6 \\
     & M$^{B}$ & \cellcolor{r2_10}0.259 & \cellcolor{r6_10}0.15 & \cellcolor{r5_10}0.243 & \cellcolor{r4_10}0.788 & \cellcolor{r5_10}2.523 & \cellcolor{r7_10}0.126 & \cellcolor{r7_10}0.11 & \cellcolor{r8_10}0.201 & \cellcolor{r7_10}0.495 & \cellcolor{r8_10}1.107 & \cellcolor{r7_10}0.176 & \cellcolor{r10_10}3.08 & \cellcolor{r8_10}0.145 & \cellcolor{r8_10}0.324 & \cellcolor{r7_10}0.686 & \cellcolor{r6_10}0.605 & \cellcolor{r10_10}3.56 & \cellcolor{r9_10}0.587 & \cellcolor{r8_10}1.010 & \cellcolor{r8_10}4.489 & \cellcolor{r6_9}13/20 & \cellcolor{r8_10}7.0 \\
     & M$^{C}$ & \cellcolor{r2_10}0.259 & \cellcolor{r9_10}2.71 & \cellcolor{r4_10}0.241 & \cellcolor{r5_10}0.813 & \cellcolor{r4_10}2.499 & \cellcolor{r7_10}0.126 & \cellcolor{r10_10}12.4 & \cellcolor{r2_10}0.138 & \cellcolor{r3_10}0.445 & \cellcolor{r6_10}1.096 & \cellcolor{r7_10}0.176 & \cellcolor{r7_10}-0.59 & \cellcolor{r5_10}0.127 & \cellcolor{r5_10}0.237 & \cellcolor{r4_10}0.507 & \cellcolor{r6_10}0.605 & \cellcolor{r9_10}3.38 & \cellcolor{r6_10}0.576 & \cellcolor{r6_10}0.991 & \cellcolor{r4_10}4.277 & \cellcolor{r4_9}15/20 & \cellcolor{r5_10}5.5 \\
     & M$^{B+C}$ & \cellcolor{r2_10}0.259 & \cellcolor{r3_10}-0.31 & \cellcolor{r2_10}0.235 & \cellcolor{r2_10}0.760 & \cellcolor{r2_10}2.429 & \cellcolor{r7_10}0.126 & \cellcolor{r5_10}-0.66 & \cellcolor{r9_10}0.202 & \cellcolor{r8_10}0.496 & \cellcolor{r8_10}1.107 & \cellcolor{r7_10}0.176 & \cellcolor{r5_10}-1.13 & \cellcolor{r5_10}0.127 & \cellcolor{r3_10}0.232 & \cellcolor{r3_10}0.493 & \cellcolor{r6_10}0.605 & \cellcolor{r4_10}0.60 & \cellcolor{r4_10}0.567 & \cellcolor{r7_10}0.992 & \cellcolor{r5_10}4.286 & \cellcolor{r3_9}17/20 & \cellcolor{r4_10}4.8 \\
     & M+H$^{B}$ & \cellcolor{r8_10}0.277 & \cellcolor{r7_10}0.77 & \cellcolor{r9_10}0.261 & \cellcolor{r8_10}0.862 & \cellcolor{r10_10}2.992 & \cellcolor{r3_10}0.122 & \cellcolor{r6_10}-0.35 & \cellcolor{r5_10}0.196 & \cellcolor{r4_10}0.455 & \cellcolor{r4_10}1.076 & \cellcolor{r3_10}0.161 & \cellcolor{r8_10}-0.57 & \cellcolor{r7_10}0.134 & \cellcolor{r7_10}0.299 & \cellcolor{r8_10}0.759 & \cellcolor{r3_10}0.602 & \cellcolor{r7_10}3.15 & \cellcolor{r8_10}0.578 & \cellcolor{r5_10}0.985 & \cellcolor{r7_10}4.456 & \cellcolor{r8_9}10/20 & \cellcolor{r7_10}6.3 \\
     & M+H$^{C}$ & \cellcolor{r8_10}0.277 & \cellcolor{r10_10}4.25 & \cellcolor{r8_10}0.260 & \cellcolor{r10_10}0.877 & \cellcolor{r9_10}2.956 & \cellcolor{r3_10}0.122 & \cellcolor{r9_10}7.11 & \cellcolor{r1_10}\textbf{0.136} & \cellcolor{r1_10}\textbf{0.420} & \cellcolor{r3_10}1.074 & \cellcolor{r3_10}0.161 & \cellcolor{r6_10}-0.94 & \cellcolor{r4_10}0.125 & \cellcolor{r6_10}0.266 & \cellcolor{r6_10}0.670 & \cellcolor{r3_10}0.602 & \cellcolor{r8_10}3.38 & \cellcolor{r5_10}0.568 & \cellcolor{r4_10}0.961 & \cellcolor{r2_10}4.255 & \cellcolor{r7_9}12/20 & \cellcolor{r5_10}5.5 \\
     & M+H$^{B+C}$ & \cellcolor{r8_10}0.277 & \cellcolor{r5_10}0.01 & \cellcolor{r7_10}0.255 & \cellcolor{r7_10}0.851 & \cellcolor{r8_10}2.904 & \cellcolor{r3_10}0.122 & \cellcolor{r3_10}-1.00 & \cellcolor{r7_10}0.197 & \cellcolor{r5_10}0.458 & \cellcolor{r5_10}1.078 & \cellcolor{r3_10}0.161 & \cellcolor{r2_10}-1.25 & \cellcolor{r3_10}0.124 & \cellcolor{r4_10}0.233 & \cellcolor{r5_10}0.573 & \cellcolor{r3_10}0.602 & \cellcolor{r3_10}0.49 & \cellcolor{r2_10}0.556 & \cellcolor{r3_10}0.960 & \cellcolor{r3_10}4.268 & \cellcolor{r5_9}14/20 & \cellcolor{r3_10}4.5 \\
     & G$^{B}$ & \cellcolor{r7_10}0.275 & \cellcolor{r2_10}-0.43 & \cellcolor{r1_10}\textbf{0.233} & \cellcolor{r3_10}0.772 & \cellcolor{r3_10}2.447 & \cellcolor{r6_10}0.123 & \cellcolor{r2_10}-1.26 & \cellcolor{r4_10}0.183 & \cellcolor{r6_10}0.463 & \cellcolor{r2_10}1.018 & \cellcolor{r2_10}0.134 & \cellcolor{r3_10}-1.22 & \cellcolor{r1_10}\textbf{0.105} & \cellcolor{r1_10}\textbf{0.176} & \cellcolor{r1_10}\textbf{0.365} & \cellcolor{r10_10}0.633 & \cellcolor{r2_10}0.33 & \cellcolor{r3_10}0.558 & \cellcolor{r1_10}\textbf{0.942} & \cellcolor{r1_10}\textbf{3.987} & \cellcolor{r2_9}18/20 & \cellcolor{r2_10}3.0 \\
     & \textbf{G+H$^{B}$} & \cellcolor{r6_10}0.265 & \cellcolor{r1_10}\textbf{-0.45} & \cellcolor{r2_10}0.235 & \cellcolor{r1_10}\textbf{0.750} & \cellcolor{r1_10}\textbf{2.238} & \cellcolor{r2_10}0.121 & \cellcolor{r1_10}\textbf{-1.30} & \cellcolor{r3_10}0.167 & \cellcolor{r2_10}0.440 & \cellcolor{r1_10}\textbf{0.972} & \cellcolor{r1_10}\textbf{0.133} & \cellcolor{r1_10}\textbf{-1.29} & \cellcolor{r2_10}0.114 & \cellcolor{r2_10}0.206 & \cellcolor{r2_10}0.386 & \cellcolor{r2_10}0.590 & \cellcolor{r1_10}\textbf{0.33} & \cellcolor{r1_10}\textbf{0.541} & \cellcolor{r2_10}0.958 & \cellcolor{r6_10}4.322 & \cellcolor{r1_9}\textbf{19/20} & \cellcolor{r1_10}\textbf{2.0} \\
    \hline
    \end{tabular*}  
    {\raggedright\scriptsize
    $^{\dagger}$ \textbf{Supervision strategy}, where \textbf{H} is the Huber loss, \textbf{G} the KL-divergence loss with error-conditioned Gaussian labels~(\ref{eq:kl}), \textbf{M} the maximum likelihood estimation with a bin mixture~\cite{lu2024rtmo}, and \textbf{N} the negative log-likelihood with a covariance decoder. \textbf{M} and \textbf{G} are also evaluated with the Huber term added (\textbf{+H}). \\
    $^{B/C/B+C}$ \textbf{Uncertainty source} used in fusion, where $^{B}$ is the bin variance~(\ref{eq:bin_variance}), $^{C}$ the covariance decoder, and $^{B+C}$ the sum of the two variances. Note that, \textbf{Reg.}, \textbf{M}, and \textbf{M+H} use a separate covariance decoder.
    \par}
    \vspace{-14pt}
\end{table*}

%% file: tables/ablation_encoding.tex
\begin{table}[t!]
    \centering
    \caption{Bin-encoding ablation on the \emph{raw network} AVE [m/s], RTE [m], and NLL [nats], varying only the bin embedding. \textbf{SPE} is our learnable sine positional encoding of~(\ref{eq:spe}).}
    \vspace{-4pt}
    \label{tab:ablation_enc}
    \renewcommand{\arraystretch}{1.15}
    \setlength{\tabcolsep}{0.9pt}
    \scriptsize
    \begin{tabular*}{\columnwidth}{@{\extracolsep{\fill}}c l | ccc | ccc | ccc | ccc}
    \hline\hline
    \multicolumn{2}{c|}{\multirow{2}{*}{Encoding}} & \multicolumn{3}{c|}{AI-IO} & \multicolumn{3}{c|}{NanoBench} & \multicolumn{3}{c|}{TII-RATM} & \multicolumn{3}{c}{NeuroBEM} \\
    \cline{3-5}\cline{6-8}\cline{9-11}\cline{12-14}
    \multicolumn{2}{c|}{} & AVE & RTE & NLL & AVE & RTE & NLL & AVE & RTE & NLL & AVE & RTE & NLL \\
    \hline\hline
     \multicolumn{2}{c|}{Reg.} & \cellcolor{r1_4}\textbf{0.263} & \cellcolor{r2_4}0.709 & \cellcolor{r4_4}-0.11 & \cellcolor{r4_4}0.158 & \cellcolor{r4_4}0.517 & \cellcolor{r4_4}-0.78 & \cellcolor{r4_4}0.206 & \cellcolor{r4_4}0.365 & \cellcolor{r2_4}-1.15 & \cellcolor{r3_4}0.607 & \cellcolor{r4_4}0.921 & \cellcolor{r4_4}0.63 \\
    \hline
    \multirow{3}{*}{\rotatebox[origin=c]{90}{VeloBins}} & Direct & \cellcolor{r4_4}0.308 & \cellcolor{r4_4}0.843 & \cellcolor{r3_4}-0.36 & \cellcolor{r2_4}0.122 & \cellcolor{r3_4}0.396 & \cellcolor{r1_4}\textbf{-1.32} & \cellcolor{r1_4}\textbf{0.123} & \cellcolor{r3_4}0.314 & \cellcolor{r3_4}-1.08 & \cellcolor{r4_4}0.610 & \cellcolor{r2_4}0.858 & \cellcolor{r2_4}0.35 \\
     & PE & \cellcolor{r2_4}0.264 & \cellcolor{r3_4}0.715 & \cellcolor{r2_4}-0.44 & \cellcolor{r3_4}0.127 & \cellcolor{r2_4}0.385 & \cellcolor{r3_4}-1.20 & \cellcolor{r3_4}0.163 & \cellcolor{r2_4}0.292 & \cellcolor{r4_4}-0.94 & \cellcolor{r1_4}\textbf{0.586} & \cellcolor{r3_4}0.860 & \cellcolor{r3_4}0.36 \\
     & \textbf{SPE} & \cellcolor{r3_4}0.265 & \cellcolor{r1_4}\textbf{0.700} & \cellcolor{r1_4}\textbf{-0.45} & \cellcolor{r1_4}\textbf{0.121} & \cellcolor{r1_4}\textbf{0.361} & \cellcolor{r2_4}-1.30 & \cellcolor{r2_4}0.133 & \cellcolor{r1_4}\textbf{0.280} & \cellcolor{r1_4}\textbf{-1.29} & \cellcolor{r2_4}0.590 & \cellcolor{r1_4}\textbf{0.835} & \cellcolor{r1_4}\textbf{0.33} \\
    \hline
    \end{tabular*}
    \vspace{-12pt}
\end{table}